# ARTIFICIAL NEURAL NETWORK BASED OPTICAL CHARACTER RECOGNITION

Vivek Shrivastava[1] and Navdeep Sharma[2]

[1]platinum012@gmail.com
[2]navdeep.sharma21@gmail.com

## ABSTRACT

*Optical Character Recognition deals in recognition and classification of characters from an image. For the recognition to be accurate, certain topological and geometrical properties are calculated, based on which a character is classified and recognized. Also, the Human psychology perceives characters by its overall shape and features such as strokes, curves, protrusions, enclosures etc. These properties, also called Features are extracted from the image by means of spatial pixel-based calculation. A collection of such features, called Vectors, help in defining a character uniquely, by means of an Artificial Neural Network that uses these Feature Vectors.*

## 1. INTRODUCTION

Automated Optical Character Recognition has gained impetus largely due to its application in the fields of Computer Vision, Intelligent Text Recognition applications and Text based decision-making systems. The approach taken to solve the OCR problem was based on psychology of the characters as perceived by the humans. Thus the geometrical features of a character and its variants were considered for recognition [1].

Later, a Template-matching approach was followed that involved comparing input characters to pre-defined templates. This method recognized characters either as an exact match or no match at all [2]. It also didn't accommodate effects like tilts and style variations that didn't involve major shape alterations.

Another approach, namely Recognition using Correlation Coefficients was based on the Cross Correlation of input characters or their transforms, with the database templates; so as to accommodate minor differences was used. It introduced False or Erroneous Recognition among characters very similar in shape, such as 'I' & 'J', 'B' & '8', 'O', 'Q' & '0' etc.

The solution to this problem lies in ANN, a system that can perceive and recognize a character based on its topological features such as shape, symmetry, closed or open areas, and number of pixels. The advantage of such a system is that it can be trained on 'samples' and then can be used to recognize characters having a similar (not exact) feature set.

The ANN used in this system gets its inputs in the form of Feature Vectors. This is to say that every feature or property is separated and assigned a numerical value. The set of these numerical values that can be used to uniquely identify each character is called its Vector. Thus, a Vector Database is utilized to train the network, so as to enable it to effectively recognize each character, based on its topological properties.

To generate the Vector Database, a set of properties or features are chosen that 'define' the character according to the human perception. To make the system generic or open to all the variants of the OCR problem, the Vector Generation step is made to be automatic in calculations and diverse enough to increase precision. A Feature is any property of the image that can be used to identify the character, such as Curves, Closed areas, Horizontal & Vertical lines, Symmetry, Contours, Projections etc [3]. The higher the number of such different features available for use, the higher is the precision of the recognition. Thus, Automated Feature Extraction is another very important aspect of the OCR problem.

Yet another dimension can be added to the OCR systems to aid it in efficient recognition. Various image transforms are available for use to the system designers, such as Fourier Transform, Discreet Cosine Transform etc. A transform based calculation hold advantage over simple pixel based calculation as a transform gives us information about various properties of the image, like frequency, noise etc. Thus, transform provide better control over the information stored in an image. One of the very useful and fast transform is the Walsh-Hadamard Transform (WHT). The Walsh-Hadamard Transform (WHT) is a suboptimal, non-sinusoidal, orthogonal transformation that decomposes a signal into a set of orthogonal, rectangular waveforms called Walsh functions. The transformation has no multipliers and is real because the amplitude of Walsh (or Hadamard) functions has only two values, +1 or -1 [4].

## 2. PROBLEM DEFINITION

A character can be written in a number of ways differing in shape and properties, such as Tilt, stroke, Cursivity and Overall shape. A plethora of Fonts are available for use in any commonly used Word Processing Application Software. Yet, while perceiving any text written in a variety of ways, humans can easily recognize and read each character. This is because the human perception processes the information by the features that define a character's shape in an overall fashion. Thus, while modeling the human perception model in machines, a rugged Feature Extraction algorithm is needed before an ANN can be applied for classification of characters.

Furthermore, OCR is aimed at developing the ability to 'read', also known as Computer Vision. There may be cases when ambiguity lies in recognition of a character or the recognized data needs to be processed as information, such as a message or signboard. In such a case, recognized data need to go for a Lexicographic lookup, probably in a dictionary or a similar document, as a form of Post-processing. Thus, a recognized character should be carefully classified, if the same 'symbol' may signify more than one character at different places.

## 3. RECOGNITION ALGORITHM

### 3.1 – Pre-processing

Any image needs some Pre-processing, before being fed to the recognition system. The first step is the conversion of any kind of image into a Binary image (the one having pixel values as '0' & '1' only). The following flowchart denotes the steps of the algorithm, 'Binarization' converts any image into a series of Black text written on a White background. Thus, it induces uniformity to all the input images. Other effects such as contrast, sharpness etc. can also be easily handled once the image has been binarized.

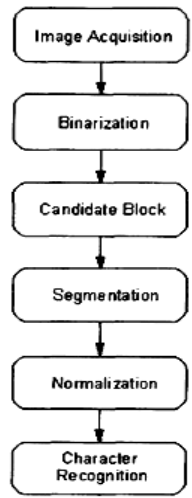

**Figure 1** – Preprocessing Flowchart

The ANN used in the system uses 'Feature Vectors' as its input. Hence, each character is segmented out from the pre-processed image. This segmentation occurs in two phases. First, each line is separated in the input image. Then each character is separated out in each line. It may be noted that the step of selecting out a 'Candidate Block' is required where only a part of image contains 'text' which needs to be recognized.

Segmentation can be done by calculating the edges of the character, where sum of 'black' pixels is zero, along the periphery of the character [3].

Then, each character so separated is normalized in terms of size and focus, so as to resemble the 'templates' that have been used for training the ANN. In this way, the input samples, processed in the same way to extract Features and generate Vectors, tend to give highly precise results.

## 3.2 – Feature Extraction

Feature Extraction serves two purposes; one is to extract properties that can identify a character uniquely. Second is to extract properties that can differentiate between similar characters.

A character can be written in a variety of ways, and yet can be easily recognized correctly by a Human. Thus, there exist a set of principles or logics that surpass all variation differences. Thus, the features used by the system work upon such properties which are close to the psychology of the characters.

A set of different types of features has been used to identify the characters, in our algorithm. These include Sum of pixels along the horizontal lines drawn at various distances along the character height as shown in Figure 2. These parameters differ from one character to another based on its width profile variation along the height [3]. Considering a binary image 'I' that contains 'm' rows and 'n' columns, having a black foreground (text) and a white background, then each pixel has a value '1' or '0' depending on whether it is white or black. So, the sum of all relevant pixels at a certain object height, say $c*m$, ($c$=scale constant, $0<c<1$), is given by,

$$Sum_{cm} = \sum_{p=1}^{n} I(c*m,p). \qquad \text{(Equation 1)}$$

where, I(c*m,p) = Black pixel at location (c*m,p).

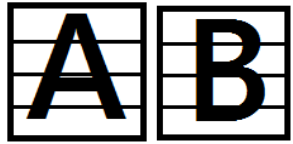

**Figure 2** – Horizontal Lines at different heights.

Similarly, a set of vertical lines drawn at various distances along the width, depicting the sum of pixels, can also serve as another feature set, as shown. Mathematically, the sum of pixels along the vertical line at a width of c*n is given by,

$$Sum_{cn} = \sum_{p=1}^{m} I(p,c*n). \qquad \text{(Equation 2)}$$

where, I(p,c*n) = Black pixel at location (p,c*n)

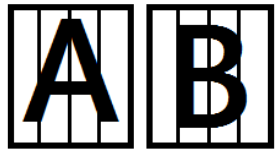

**Figure 3** – Vertical Lines along the width.

Symmetry is another parameter that can be used to reduce ambiguity among characters, such as '8' and 'B' can be differentiated based on its Horizontal Symmetry while 'I' and 'J' can be differentiated easily based on their Vertical Symmetry. It should be noted that these parameters show the 'Degree of symmetry', i.e. a decimal value between 0 (No symmetry) to 1 (Perfect symmetry), rather than 'True' or 'False'. For this, we create a matrix, say M having the first half (horizontal or vertical) part to be the mirror image of the second half. Then, the correlation is found between 'M' and 'I'. This level of correlation gives us the amount of symmetry the character has.

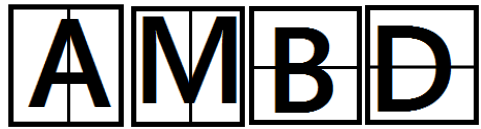

**Figure 4** – Horizontal & Vertical Symmetry in characters.

Another paradigm of character recognition is the number of closed areas in its shape. Characters such as 'A','P','D' and 'Q' have one closed area, while others such as 'B' and '8' have two. There also exist characters which are open, such as 'H', '7', 'C' etc. This parameter also serves to broadly classify characters based on its openness or closeness.

Comparison can be done in yet another form, namely, using the WHT coefficients. A database was created that contained the Walsh-Hadamard Transforms of each image 'template'. Thus, each character's value is available in terms of WHT coefficients. By calculating the WHT equivalent of the input character, and calculating its correlation to each sample in the WHT database, so maintained, we are able to find the character having the highest similarity to the input character. Thus, classification is done using transformed structure, rather than pixel-based arithmetic.

Figure 5 shows the WHT coefficients of the character 'A'. It shows the variation in the magnitude of the WHT coefficients, along the Sequency index of the coefficients. Although the whole range of coefficients was utilized for comparison, higher order coefficients (higher than 40, in the figure), having very small magnitudes can be neglected for more complex databases.

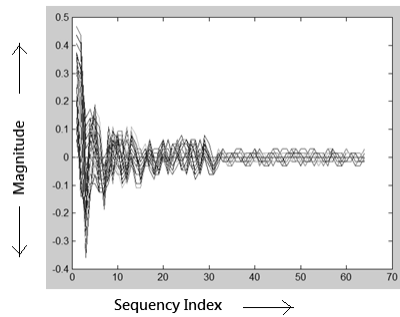

**Figure 5** – Variation in magnitude of the WHT coefficients of the character 'A'.

Another paradigm of character recognition is the number of closed areas in its shape. Characters such as 'A','P','D' and 'Q' have one closed area, while others such as 'B' and '8' have two.

The main idea behind calculating a number of different parameters is to increase the differences among the characters, so as to make the recognition easier. Thus, addition of parameters tends to increase the entropy of the system. Hence another parameter that was used is 'Sum of values of the other parameters'. It should be noted that while others were primary features, i.e. calculated directly from the images, this one is a secondary parameter, i.e. calculated from primary feature values.

Thus, the eleven parameters used for training the ANN are :

1.  H30: Sum of pixels at 30% character height as given by $Sum_{cm}$ as calculated using Equation 1, with c=0.3, m=42 and n=24.

2.  H50: Sum of pixels at 50% character height as given by $Sum_{cm}$ as calculated using Equation 1, with c=0.5, m=42 and n=24.

3.  H80: Sum of pixels at 80% character height as given by $Sum_{cm}$ as calculated using Equation 1, with c=0.8, m=42 and n=24.

4.  V30: Sum of pixels at 30% character width as given by $Sum_{cn}$ as calculated using Equation 2, with c=0.3, m=42 and n=24.

5.  V50: Sum of pixels at 50% character width as given by $Sum_{cn}$ as calculated using Equation 2, with c=0.5, m=42 and n=24.

6.  V80: Sum of pixels at 80% character width as given by $Sum_{cn}$ as calculated using Equation 2, with c=0.8, m=42 and n=24.

7. Hsym: Correlation between an input character sample 'I' with 'X'. It should be noted that 'X' is generated from 'I' where left half of 'X' is same as that of 'I' and right half of 'X' is the mirror image of its left half.

8. Vsym: Correlation between an input character sample 'I' with 'X'. It should be noted that 'X' is generated from 'I' where upper half of 'X' is same as that of 'I' and lower half of 'X' is the mirror image of its upper half.

9. Pos: It denotes the 'position' at which the calculated WHT of an input sample shows highest correlation to the WHT database, containing Walsh Hadamard Transform of each character in order A-Z and 1-0.

10. CC: It gives a measure of the number of closed areas in a character. In an image, all the connected pixels are given same labels. Thus if there are two sets of such pixels, as in '8', they would be labeled as '1' and '2'. Thus the highest value of 'label' gives an idea of closed areas present in the character.

11. Sumt: This is a secondary feature, calculated by adding the values of all the ten features evaluated previously. This is an additional feature, used to increase the entropy of the system, for better recognition.

## 3.3 – ANN Training and Classification

Before the character recognition can take place, the ANN is 'trained', so that it can develop the capability of mapping various inputs to the required outputs and effectively classify various characters. For training the ANN, we use the 'Vectors' generated by the 'Database Templates' using the above mentioned Feature Extraction techniques. The above mentioned 7 different types of features have been used to generate 11 parameters (some of same type but different values), which are fed to the ANN. Thus, a matrix of 11x36 values is fed to the ANN to receive 36 different values at the output, one for each character in the database.

It may be noted that the ANN uses Backpropagation algorithm for Learning. The 'Target' values are specified by the system programmer to accommodate for small recognition errors, which may be changed from application to application.

The ANN was trained for 1000 iterations, which took around 21 seconds to complete. The Training Function was set to use 'Sum Squared Error' rather than 'Mean Squared Error', because the system needed to calculate the effect of joint errors in all the parameters, rather than overall error. An error goal of 0.0001 or 0.01% was achieved by the ANN.

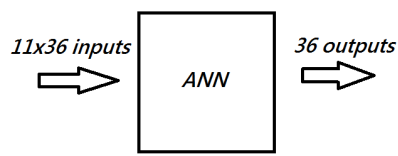

**Figure 6 –** Input-Output parameters of ANN during Training.

## 4. EXPERIMENTAL RESULTS & SUMMARY

The ANN system, being trained using standard templates of the capital alphabets and numbers (as they are); shows 100% recognition for the set of data, it was trained upon.

The system was tested upon characters from untrained fonts. It shows a recognition rate of 85.83% of 10 simple fonts, available for use, i.e. it recognizes 309 characters correctly out of 360.

Taking into consideration 10 other fonts which fall under a category of Low styled to Medium styled, the recognition rate drops to 75%, i.e. the system efficiency drops to recognizing 540 characters correctly out of 720.

This surge in performance results from false recognitions due to similarities between ambiguous characters such as I & J, K & R, O & Q, B & 8 etc. Also to be noted is the fact that as the level of styling or 'Cursivity' of the alphabets increase, ambiguity level among characters becomes more prominent.

Table 1 - Recognition Rate for Untrained Fonts by the ANN system.

| S.NO. | NAME OF THE FONT | RECOGNITION RATE |
|---|---|---|
| 1 | ARIAL | 97.20 |
| 2 | MICROSOFT SANS SERIF | 91.67 |
| 3 | CENTURY GOTHIC | 88.89 |
| 4 | TREBUCHET MS | 86.11 |
| 5 | CALIBRI | 83.33 |
| 6 | COPPERPLATE GOTHIC | 83.33 |
| 7 | LUCIDA CONSOLE | 83.33 |
| 8 | LUCIDA SANS TYPEWRITER | 83.33 |
| 9 | VERDANA | 83.33 |
| 10 | LUCIDA SANS | 77.78 |
| 11 | SEGOE UI | 75.00 |
| 12 | FRANKLIN GOTHIC BOOK | 72.22 |
| 13 | TEMPUS SANS ITC | 69.40 |
| 14 | TIMES NEW ROMAN | 69.40 |
| 15 | CANDARA | 66.67 |
| 16 | TAHOMA | 66.67 |
| 17 | BOOKMAN OLD STYLE | 63.89 |
| 18 | OCR A | 63.89 |
| 19 | COMIC SANS MS | 58.33 |
| 20 | GEORGIA | 33.33 |

The Recognition Rate table shown in Figure 7 also shows the Fonts in decreasing order of recognition. Arial performs best and recognizes all characters (35/36) except 'I'. Microsoft Sans Serif performs second best, another font where there is no error in recognizing numbers. Highest False Recognition cases appear in 'I', 'O', 'Q', '1' and '3' in that order.

# 5. CONCLUSION

The ANN based system has shown promising results due to the fact that despite being trained only on a single set of templates (independent of any pre-defined font), it not only gets trained in 21 seconds, but also can recognize the fonts (for which it was not trained) with high efficiency, as already observed.

The system has its advantages such as Less Time Complexity, Very Small Database and High Adaptability to untrained inputs, with only a small number of features to calculate as compared to the method followed in [3]. Yet, the system has a large scope for further developments.

System performance can be increased further by:

1) Increasing the DATABASE used for training the ANN, so as to enable it to recognize stylized fonts also. 2) Using better algorithms for training the ANN, so as to decrease the Time complexity while handling larger databases. 3) Better Feature Extraction techniques so as to increase the precision of results. 4) Introducing algorithms which recognize characters on the psychology or property of a character, not by comparing them to a list of templates as such. This would be the final step to induce recognition through ANN, as modelled on the Human perception of text.

## REFERENCES

[1] Ching Y. Suen and Robert J. Shillman, "Low Error Rate Optical Character Recognition of Unconstrained Handprinted Letters Based on a Model of Human Perception", *IEEE Transactions on Systems, Man, and Cybernetics, June 1977.*

[2] Vinod Chandra and R. Sudhakar, "Recent Developments in Artificial Neural Network Based Character Recognition: A Performance Study", *IEEE, 1988.*

[3] Evelina Maria De Almeida Neves, Adilson Gonzaga, Annie France Frere Slaets, "A Multi-Font Character Recognition Based on its Fundamental Features by Artificial Neural Networks", *IEEE, 1997.*

[4] D. Sasikala, R. Neelaveni, "Correlation Coefficient Measure of Multimodal Brain Image Registration using Fast Walsh Hadamard Transform", Journal of Theoretical and Applied Information Technology, 2005.